\documentclass[letterpaper, 10 pt, conference]{ieeeconf}  

\IEEEoverridecommandlockouts                              

\usepackage{times}
\usepackage{epsfig}
\usepackage{graphicx}
\usepackage{amsmath}
\usepackage{amssymb}

\usepackage[10pt]{moresize}

\usepackage[utf8]{inputenc} 
\usepackage[T1]{fontenc}    
\usepackage{url}            
\usepackage{booktabs}       
\usepackage{amsfonts}       
\usepackage{nicefrac}       
\usepackage{microtype}      
\usepackage{graphicx}
\usepackage{enumerate}
\usepackage{algpseudocode,algorithm,algorithmicx}  

\usepackage[pagebackref=true,breaklinks=true,letterpaper=true,colorlinks,bookmarks=false]{hyperref}
\usepackage{caption}

\usepackage{gensymb}
\usepackage{subcaption}
\usepackage{color,soul}
\usepackage[nomargin,inline,marginclue,draft]{fixme}

\DeclareMathOperator{\E}{\mathbb{E}}
\DeclareMathOperator{\N}{\mathcal{N}}

\newcommand{\rulesep}{\unskip\ \vrule width 0.75pt height 55px\ }

\newcommand\blfootnote[1]{%
  \begingroup
  \renewcommand\thefootnote{}\footnote{#1}%
  \addtocounter{footnote}{-1}%
  \endgroup
}

\title{\LARGE \bf Learning Real-World Robot Policies by Dreaming}

\author{
  AJ Piergiovanni \hspace{1cm} Alan Wu \hspace{1cm} Michael S. Ryoo \\
  School of Informatics, Computing, and Engineering\\
  Indiana University Bloomington\\
  \texttt{\{ajpiergi, alanwu, mryoo\}@indiana.edu} \\
}

\begin{document}

\maketitle
\thispagestyle{empty}
\pagestyle{empty}

\begin{abstract}
Learning to control robots directly based on images is a primary challenge in robotics. However, many existing reinforcement learning approaches require iteratively obtaining millions of robot samples to learn a policy, which can take significant time. In this paper, we focus on learning a realistic world model capturing the dynamics of scene changes conditioned on robot actions. Our \emph{dreaming model} can emulate samples equivalent to a sequence of images from the actual environment, technically by learning an action-conditioned future representation/scene regressor. This allows the agent to learn action policies (i.e., visuomotor policies) by interacting with the dreaming model rather than the real-world. We experimentally confirm that our dreaming model enables robot learning of policies that transfer to the real-world.
\end{abstract}

\blfootnote{This work was supported by Institute for Information \& communications Technology Promotion (IITP) grant funded by the Korean government (MSIT) (No. 2018-0-00205, Development of Core Technology of Robot Task-Intelligence for Improvement of Labor Condition).}

\section{Introduction}
\blfootnote{Example dreams: \href{https://piergiaj.github.io/robot-dreaming-policy/}{https://piergiaj.github.io/robot-dreaming-policy/}}

Learning to control robots directly based on camera input (i.e., images) has been one of the primary challenges in robotics. Recently, significant progress has been made in deep reinforcement learning (RL), enabling learning of control policies using raw image data from physics engines and video game environments \cite{mnih2013playing,heess2015learning}. These methods have also been applied to real-world robots, obtaining promising results \cite{finn2017deep,wahlstrom2015pixels}. Learning state representations with convolutional neural networks (CNNs) instead of using handcrafted states particularly benefits robots, which are required to process high dimensional image data with rich scene information.



However, many of these reinforcement learning methods require obtaining millions of samples (i.e., trajectories) over multiple training iterations to converge to an optimal policy. 
This is often very expensive for real-world robots; obtaining each sample of a robot trial may take multiple minutes even with a well-prepared setup, resulting in the total robot learning time to be weeks/months every time the setting changes or a new task is needed (e.g., 3000 robot hours in \cite{levine2016learning} and 700 hours in \cite{pinto2016supersizing}). Further, there are also safety issues when obtaining real-world robot samples. This difficulty made deep reinforcement learning works applied to simulators (e.g., MuJoCo) or computer games (e.g., Atari) much more common, in contrast to real-world robots.





In this paper, we focus on the problem of learning a \emph{dreaming model} that can emulate realistic training samples equivalent to sequences of image frames from the actual environment. Our dreaming model is a learned `world model' that also allows random sampling of initial robot images. The objective is to make the agent (i.e., a robot) learn visuomotor policies by interacting with the dreaming model instead of the real-world. Our dreaming model is learned solely based on relatively few initial videos of random robot actions. It is explicitly designed to allow learning of a good state representation space that generates realistic samples from scratch while also correctly modeling their dynamics over a series of actions.
The dreaming model synthesize artificial robot experience by generating `dreamed trajectories/samples' composed of state-action pairs, which essentially is a sequence of CNN representations.
This not only makes the policy learning much more efficient by minimizing real robot executions, but also is crucial for the situations where robot interactions with its real-world environment is expensive/prohibited.



We introduce a fully convolutional network architecture combining a variational autoencoder and an action-conditioned future regressor to implement our dreaming model.
What we mean by an `action' in this paper is a control command to the robot (e.g., go 30cm forward), and our action-conditioned future regressor models how the scene representation (i.e., the state) changes as the robot moves according to the command.
The use of the dreaming model can be viewed as an extension of previous model-based reinforcement learning works using `imagined trajectories' in addition to real trajectories~\cite{sutton1990integrated,heess2015learning,weber2017imagination}. The main difference is that (i) our state representation space is learned in a way that is optimized for generating dreamed trajectories from scratch.
The state space our dreaming model learns is explicitly trained to make any of its samples correspond to a video frame, and this allows generation of dreamed trajectories with (good) random start states. 
Further, (ii) our proposed approach attempts to learn such model from random initial trajectories, different from previous works. This could be interpreted as zero-real-trial reinforcement learning, and (iii) we confirm such capability experimentally with a real-world robot receiving real-world image frames.




Our experiments were conducted with a real-time robot in a real-world office environment. We show that learning our dreaming model based on initial random robot samples is possible, and confirm that it allows multiple different reinforcement learning algorithms to obtain real robot action policies. The learned policy directly controls the robot based on its RGB input without any further information. We also experimentally confirm that our dreaming model can be extended to learn a policy that could handle `unseen' object targets by learning a general state representation space. 

\section{Related works}
Convolutional neural networks (CNNs) have been very successful on many traditional computer vision tasks. These include image classification~\cite{krizhevsky2012imagenet}, object detection~\cite{liu2016ssd} and activity recognition in videos~\cite{carreira2017quo}. Learning features and representations optimized for the training data using CNNs enabled better recognition in these tasks.


Several works studied using CNNs for reinforcement learning of policies, designing deep reinforcement learning models. These works learn state representations from raw images, and allow control of the agents directly based on image inputs \cite{mnih2013playing,koutnik2014online,zhang2015towards,finn2016auto}. Zhang et al.~\cite{zhang2015towards} showed that policy learning of a robot is possible with synthetic images but found that the learned policy does not work well when directly applied to real-world images. Finn et al.~\cite{finn2016auto} trained an autoencoder to learn state representations to be used for policy learning of real-world robots. See \cite{lesort2018state} and \cite{bohmer2015autonomous_short} 
for a more detailed review. 

There also are model-based reinforcement learning works with state-transition models \cite{watter2015embed,lenz2015deepmpc,finn2017deep}.
Finn et al.~\cite{finn2016unsupervised} learned a LSTM model to predict future video frames based on actions, which was applied to the planning-based robot control \cite{finn2017deep}, focusing on model predictive control (MPC) rather than learning the policy function, analogous to \cite{wahlstrom2015pixels}. Embed-to-control (E2C) learned a state-transition model that was able to use control/planning algorithms to find an optimal action sequence to complete the trial, when given a start and end state (for each trial) \cite{watter2015embed}.
Vision researchers also studied predictive models, learning to anticipate future human locations \cite{kitani12} or image frames \cite{visualdynamics16}, 
but they were not learning an action-conditioned model for robot training. Rhinehart and Kitani \cite{rhinehart2017iccv} used inverse reinforcement learning to forecast the goals of human camera wearers, but it was not for visuo-motor policy learning. Pathak et al. \cite{pathak2018zero} took advantage of a state-transition model, but do not have the ability to generate trajectories starting from unseen states and always requires the goal image as an input.

Combining model-free reinforcement learning with model-based methods has also been studied. They often learn the state-transition model (as a neural network) and use the model to initialize/benefit the learning of model-free policies \cite{nagabandi2017neural,pong2018temporal}. This is also relevant to the use of `imagined trajectories' generated by applying the state-transition model to states from on-policy real-world trials. This includes Dyna \cite{sutton1990integrated} as well as more recent works \cite{gu2016continuous,venkatraman2016improved,silver2016predictron}. There are also works using state-transition models to combine reinforcement learning with planning, such as VPN~\cite{oh2017value} and I2A~\cite{weber2017imagination}. However, making them work with a real-world robot obtaining real-world image frames have been a challenging problem, and doing so without any real-on-policy-trials (i.e., solely based on initial random action samples) has not been addressed previously.
A concurrent research work \cite{ha2018world} discusses a similar idea to our research, but it was done with simulators instead of robots receiving real-world image/videos.


Other works have explored transferring policies learned in simulated environments to real environments for robots 
\cite{tzeng2015towards, christiano2016transfer,tobin2017domain}; however, these environment simulators are handcrafted rather than learned from data. Our dreaming model may be viewed as `learning' a simulator capturing real-world scene changes (from random initial samples), and using the learned simulator to learn realistic policies.




\section{Background}
We follow the standard reinforcement learning setting and formulate policy learning as learning in a Markov decision process (MDP). The MDP, $(\mathcal{S},\mathcal{A},f,r)$ consists of a state space, $\mathcal{S}$ and an action space $\mathcal{A}$. $f$ is the state transition function $f : \mathcal{S}\times\mathcal{A}\mapsto \mathcal{S}$, which maps from a state $s_t$ and action $a_t$ to its next state $s_{t+1}$. After each action, the environment gives a reward, $\mathcal{R}:\mathcal{S}\mapsto r$. We denote a trajectory (i.e., a sequence of states and actions) as $\tau=(s_0,a_0,s_1,a_1,\ldots, s_T)$ and we obtain a trajectory  $\tau=\rho(\pi)$ by sampling from a stochastic policy providing the actions: $\pi(a_t|s_t)$. The objective of reinforcement learning is to learn the policy maximizing the expected 
($\gamma$-discounted) reward from the start distribution, $J(\pi) = \E_{(s_t,a_t)\sim\rho(\pi)} [\sum_{i=0}^T \gamma^{i} \mathcal{R}(s_i)]$.

In this work, we consider continuous state and action spaces. We particularly take advantage of policy gradient methods including an actor-critic that learns an action-value $Q$ function $Q_\pi (s_t,a_t) = \E_{(s_t,a_t)\sim\rho(\pi)} [\sum_{i=t}^T \gamma^{i-t} \mathcal{R}(s_i)]$ together with the actor network $\pi(a_t | s_t)$.
Our agent (i.e., a robot) in a real-world environment receives an image input $I_t$ at every time step $t$, and we learn the function from $I_t$ to the state representation $s_t$ together with its policy. These functions are implemented in terms of CNNs, as was done in previous deep reinforcement learning works.

\section{Dreaming model}

Our dreaming model is a combination of a convolutional autoencoder and an action-conditioned future representation regressor. It enables encoding of an image frame (i.e., robot observation) into the representation abstracting scene (i.e., robot state), and allows predicting the expected future representation given any state and a robot action to take (i.e., transition).

\begin{figure*}
    \centering
      \includegraphics[width=0.9\textwidth]{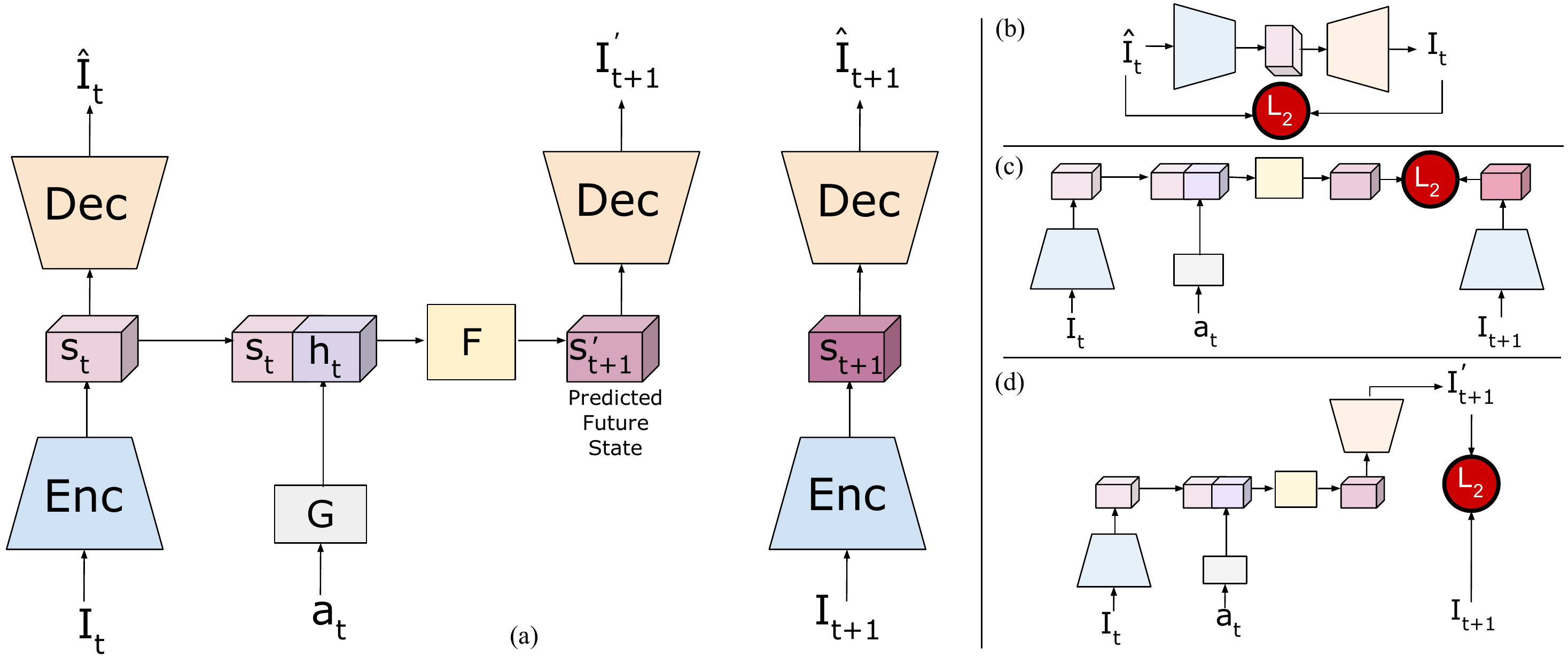}
      \caption{Illustration of our dreaming model. \textbf{(a)} The encoder, action representation, future regressor and decoder modules. \textbf{(b)} Image reconstruction loss for the autoencoder. \textbf{(c)} $L_2$ loss for the future regressor. \textbf{(d)} Future image reconstruction loss for the future image prediction. Rectangles in the figure are the CNN layers, and 3-D cuboids are the representations. Circles indicate losses.}
      \label{fig:dream-model}
\end{figure*}




More specifically, our dreaming model consists of the following four function components, governed by the learned parameters $\theta_{Enc}$, $\theta_{Dec}$, $\theta_f$, and $\theta_R$:
\begin{align*}
\mbox{\textbf{State Encoder} } & Enc_{\theta}: I \mapsto s
\\
\mbox{\textbf{State Decoder} } & Dec_{\theta}: s \mapsto I
\\
\mbox{\textbf{State-transition} } & f_{\theta}: s_t,a_t \mapsto s_{t+1}
\\
\mbox{\textbf{Reward/End Detector} } & R_{\theta}: s \mapsto r
\end{align*}
$I_t = Dec(s_t = Enc(I_t))$ is a variational autoencoder that learns a compact scene representation, to be used as the state: $s_t$.
$s_{t+1} = f(s_t, a_t)$ is a state transition model which is also often described as `future representation regression' in computer vision.
$r_t = R(s_t)$ is a CNN trained to predict the immediate reward corresponding to the state $s_t$. Note that this reward is often sparse, being constant in all states other than a terminal state.
All these functions are fully convolutional neural networks.

The important properties of our dreaming model are that (1) its state space is learned to make decoding of current/future robot states back to real-world image frames possible and that (2) the space is explicitly trained to make sense even when we randomly sample states from it. Having such realistic model enables our robot to learn its action policy by generating `dreamed trajectories' from any random starting state/image, unlike prior works using imagined trajectories.

The full model is shown in Fig.~\ref{fig:dream-model}, together with the description of our losses to train the entire dreaming model. Note that none of our losses are dependent on the policy to be learned: we are able to learn the dreaming model from trajectories of a random initial policy.

\subsection{Learning state representation}


We learn a state representation based on a variational autoencoder (VAE) \cite{kingma2014auto}
which takes an image as input and learns a latent state representation optimized to reconstruct the given image. Unlike previous works using VAEs, we maintain spatial information by designing a `convolutional' VAE, where our latent representation is a $W\times H\times C$ tensor rather than a $C$-dimensional vector in traditional VAEs. Here, $W$ is the spatial width of the representation, $H$ is the height, and $C$ is the number of channels. VAEs assume a prior distribution over the latent space $s \sim \N(0,1)$ and minimize the KL-divergence between the latent encoded representation, $Enc(I) \sim q(s|I)$ and the prior. This is an important property allowing us to randomly sample the latent space $s\sim \N(0,1)$ to obtain a new state that corresponds to a realistic image, once learned. The encoder outputs $\mu,\sigma = Enc(I)$ and we sample from $s\sim\N (\mu,\sigma)$ to obtain the state representation during the training. Once the training of the autoencoder is complete, we set $s=\mu$.

Given the encoder CNN $Enc$ and decoder CNN $Dec$, we minimize the following loss (Fig.~\ref{fig:dream-model} (b)):
\begin{equation}
    \label{eq:vae}
    \mathcal{L}_{VAE} = ||Dec(Enc(I)) - I||_2 + D_{KL}(q(s|I)||p(s))
\end{equation}
The state representation model is responsible for abstracting the important scene information into a small, latent state representation. 

\subsection{Learning the state-transition model}

\begin{figure*}
\centering
\begin{minipage}{0.48\textwidth}
\begin{algorithm}[H]
  \caption{Actor-Critic policy learning in a dream
    \label{alg:policy-learning}}  
  \begin{algorithmic}  
    \Function{Actor-Critic}{$f, R$}
    \State Initialize policy $\pi_\theta$, value function $V_\phi$
    \For{$i=0$ to num episodes}
    \State Get random starting state, $s_0 \sim \N(0,1)$
    \For{$t=0$ to $T$}
        \State $a_t \sim \pi_\theta(a_t|s_t)$
        \State $s_{t+1} = f(s_t, a_t)$
        \State $y = R(s_t) + \gamma V_\phi(s_{t+1})$
        \State $\hat{A} = y - V_\phi(s_t)$
        \State $\nabla_\theta J(\theta) \approx \nabla_\theta \log \pi_\theta (a_t|s_t) \hat{A}$
        \State $\theta = \theta + \alpha\nabla_\theta J(\theta)$
        \State $\phi = \phi + \nabla_\phi |V_\phi(s_t) - y|$
    \EndFor
    \EndFor
    \EndFunction
  \end{algorithmic}  
\end{algorithm}
\end{minipage}
\hfill
\begin{minipage}{0.48\textwidth}
\centering
\includegraphics[width=0.68\linewidth]{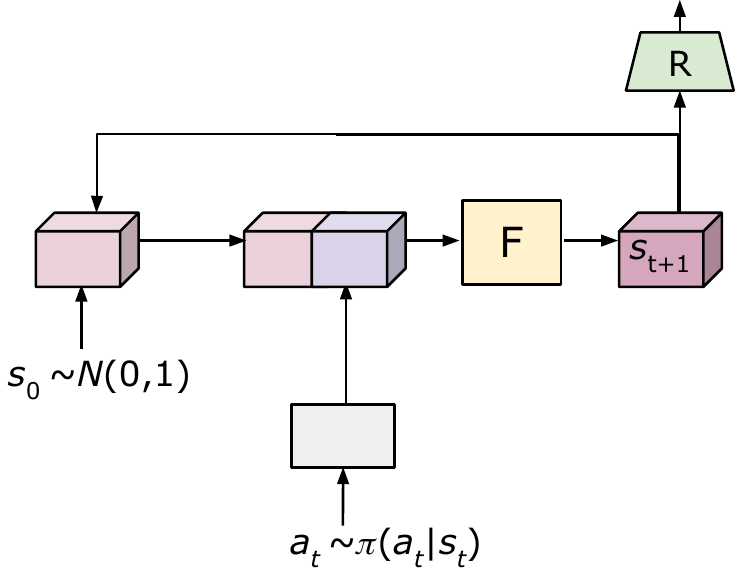}
\caption{Illustration of how our dreaming works for the policy learning. A random start state is sampled and then actions are sampled from the policy. The state-transition model predicts the next state. This process is repeated until the end detector signals the end of the trial.}
\label{fig:test}
\end{minipage}
\vspace{-18pt}
\end{figure*}

We learn a state-transition model by extending the future regression formulation from~\cite{lee2017learning} to make it action-conditioned. Our state-transition model takes the current state, $s_t$, and action, $a_t$, as input and outputs the next state, $s_{t+1}$. Since our state representation is convolutional, we train a CNN to transform the action vector $a$ also into a convolutional representation by using a fully connected layer followed by depooling to create a $W\times H\times C'$ representation, which is again followed by several convolution layers to learn a convolutional action representation:
\begin{equation}
    G(a) = conv(conv(depool(fc(a))))
\end{equation}
where $conv$ is a convolutional layer, $depool$ is the spatial depooling layer, and $fc$ is the fully connected layer.
We concatenate the action representation with the state representation along the channel axis, and use several fully convolutional layers to predict/regress the next state:
\begin{equation}
    s_{t+1} = f(s_t, a_t) = F([s_t, G(a_t)])
\end{equation}

Our method of applying spatial depooling and convolutional layers to the action vector makes preserving spatial information in the state representation possible.
A previous CNN for robot action learning (e.g., \cite{finn2017deep}) or state-transition model learning (e.g., \cite{ha2018world}) represents its state as a vector (i.e., 1-D tensor), so that it can be directly concatenated with the action vector.
On the other hand, our CNN architecture represents its state-action pair as a full 3-D tensor to better preserve spatial information in it, thereby performing superior to the previous use of linear state-action representations. (We confirm this experimentally in Section \ref{sec:exp}).

We combine the state-transition model with the state representation model to obtain the predicted future image: $I'_{t+1} = Dec(f(Enc(I_t), a_t))$. 
Then, we formulate our loss as follows:
\begin{equation}
\begin{split}
        \mathcal{L}_{f} &= ||f(Enc(I_t), a_t) - Enc(I_{t+1})||_2 \\
        &+ ||Dec(f(Enc(I_t), a_t)) - I_{t+1}||_2
\end{split}
\end{equation}
Using this loss enables our learning to minimize the difference between the the future-regressed state and the true future state + the difference between the true next image and predicted future image. This is the combination of Fig.~\ref{fig:dream-model} (c) and (d). 

We jointly train the state-transition model and the representation by minimizing the following loss:
\begin{equation}
\mathcal{L} = \mathcal{L}_{VAE} + \lambda \mathcal{L}_f    
\end{equation}
where $\lambda=0.5$ is the weight for combining the losses. This essentially is the combination of the losses described in Fig.~\ref{fig:dream-model} (b), (c), and (d).



\section{Policy learning in a dream}

Given the learned state representation space, $\mathcal{S}$, trained state-transition model, $f$, and sparse reward/end detector CNN, $R$, we can learn a policy, $\pi$, to maximize the expected reward using any reinforcement learning algorithm.
Since our state representation is learned using a variational autoencoder, we are able to start a trajectory from an unseen state by randomly sampling (i.e., generating) from the prior, $s_0\sim\N(0,1)$.
We then obtain a `dreamed' trajectory $\tau = (s_0, a_0\sim\pi(a_0|s_0), s_1=f(s_0, a_0), a_1\sim\pi(a_1|s_1), s_2=f(s_1, a_1), \ldots)$ by following our action policy and transition model. 
We compute the discounted reward for the dreamed trajectory as $\sum_{t=0}^T \gamma^{T-t} R(s_t)$, and update the policy being learned accordingly. Note that we can also start from seen states using the decoder and an image to obtain $s_0=Enc(I)$ where $I$ is from the real robot sample.

In Algorithm~\ref{alg:policy-learning}, we show an example of actor-critic policy learning in a dream.
This allows the agent to `dream' unseen starting states and explore the environment without any real-world trials. Our approach is able to benefit any reinforcement policy learning methods in general, and we are showing an actor-critic method as an example. Fig.~\ref{fig:test} illustrates how our model generates dreamed trajectories.

\section{Experiments}
\label{sec:exp}
\begin{figure}
    \centering
      \includegraphics[width=0.47\textwidth]{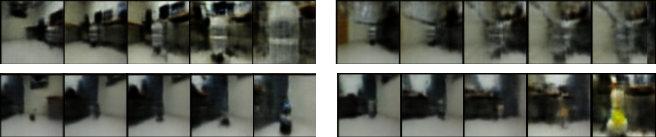}
      \caption{Four dreamed trajectory examples from random start states. For instance, approaching the air filter object is dreamed in the top-left example. Leftmost images of each sequence are the (decoded) randomly generated start states. Although the images look realistic, none of these are from real robot samples.
}
      \label{fig:rand-dream}
\end{figure}

\begin{figure}
    \centering
      \includegraphics[width=0.47\textwidth]{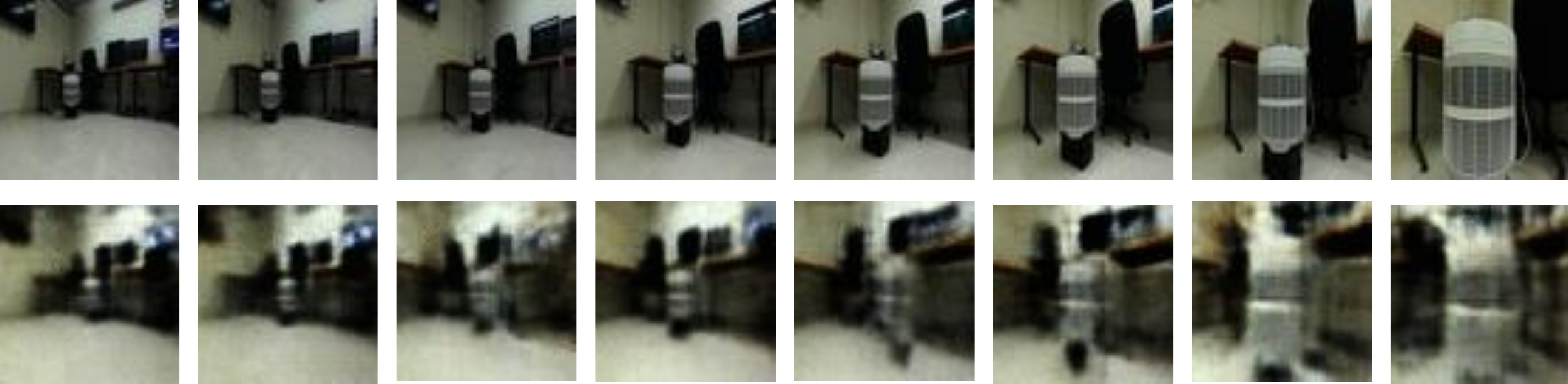}
      \caption{Comparison of (top) a real trajectory and (bottom) a dreamed trajectory. For this figure, the dreamed trajectory was given the start image and the actions were the same as the real trajectory. The dreamed images look realistic and match the true trajectory.}
      \label{fig:real-v-dream}
\end{figure}

\subsection{Environment}
We use a complex real-world office room as our environment and have a ground mobility robot, TurtleBot 2, interacting with a target object. Our environment consists of 7 different target objects such as a volleyball, shopping bag, and backpack. The robot only obtains its camera input, which is the first-person view of the room. In each trial, we vary the initial location of the robot and the target object location.

We represent actions of our ground mobility robot in polar coordinates, $a_t = (\Delta\alpha, \Delta v, \Delta\omega)$, where $\Delta\alpha$ controls the initial rotation, $\Delta v$ determines how much to move and $\Delta\omega$ determines the final rotation. We constrain the continuous action space so that $\Delta\alpha,\Delta\omega\in [-30, 30]\degree$ and $\Delta v\in [0,20]$cm.

To train the dreaming model, we collect a dataset consisting of 40,000 images (400 random trajectories) with the various target objects in different locations in the room. We allow the robot to take random actions in the environment and we store the starting image, action and resulting image pairs as $(I_t, a_t, I_{t+1})$. We use this data to train our dreaming model: the autoencoder and future regression model. 
We then annotated the images corresponding to the goal state to train the reward/end detector CNN. This CNN determines when the goal state is reached.

We emphasize once more that no real robot samples were provided except initial random action policy samples in all our experiments. Even without such real on-policy samples, our approach learns to control the robot directly from RGB frames (i.e., visuomotor policy learning). This is done without any localization, scene reconstruction, or object detection.

\subsection{Baselines and dreaming models}

In order to confirm the effectiveness of our dreaming model (which is the combination of the jointly learned convolutional representation encoder and state-transition model), we implemented several baseline models that can also generate imagined trajectories.


We implemented the standard approach of learning a state-transition model on top of the linear CNN-based state representations, as was done in \cite{heess2015learning,oh2017value,weber2017imagination}. Linear CNN-based state representations have been widely used (e.g., \cite{mnih2013playing}), and the above works learn such representations jointly with the state-transition model.
These approaches do not use a decoder and are usually trained using on-policy trials. To obtain a state representation based on these approaches in our scenario of only using random initial trajectories for the learning, we compare two different variations: (1)
a CNN initialized with ImageNet training, and (2) a CNN trained to detect the goal state. In both these baselines, we use the output from the final fully connected layer before the classification as the state representation. We also compare to (3) E2C \cite{watter2015embed} and (4) a dreaming model using the method in E2C to learn state-transition model. Note that E2C requires both the start and end state for each trial (i.e., more data than we provide the other models) for it to function.
We also implemented (5) a version of our dreaming model using a variational autoencoder to learn a `linear' state representation similar to \cite{ha2018world}.
Finally, we implemented (6) the full version of our dreaming model with convolutional state and action representations. We describe our architectures and training details in Appendix A (in the supplementary material).



\paragraph{Evaluating accuracies of the learned dreaming models}
As our initial experiment, we directly compare the normalized $L_1$ distance (Eq.~\ref{eq:rel}) between predicted vs. actual future state representations in Table~\ref{tab:norml1}. This captures the relative difference between the future predicted state and the true future state.
The (normalized) distance metrics we used are as follows:
\begin{equation}
\begin{split}
    \label{eq:rel}
    D_1(s_t,a_t, s_{t+1}, f) &= \frac{||f(s_t, a_t) - s_{t+1}||}{||s_t - s_{t+1}||}\\
    D_2(s_t,a_t, s_{t+1}, f) &= \frac{||f(s_t, a_t) - s_{t+1}||}{\frac{1}{N}\sum(||s_i - s_j||)}, \forall i,j
    \end{split}
\end{equation}
We find that the use of a decoder in our dreaming model greatly improves the accuracy over the conventional models and that the state-transition model learned by E2C is not great for our task. We also confirm that using our convolutional state/action representation design provides the best accuracy.

\begin{figure}
    \centering
    \centering
      \includegraphics[width=0.85\linewidth]{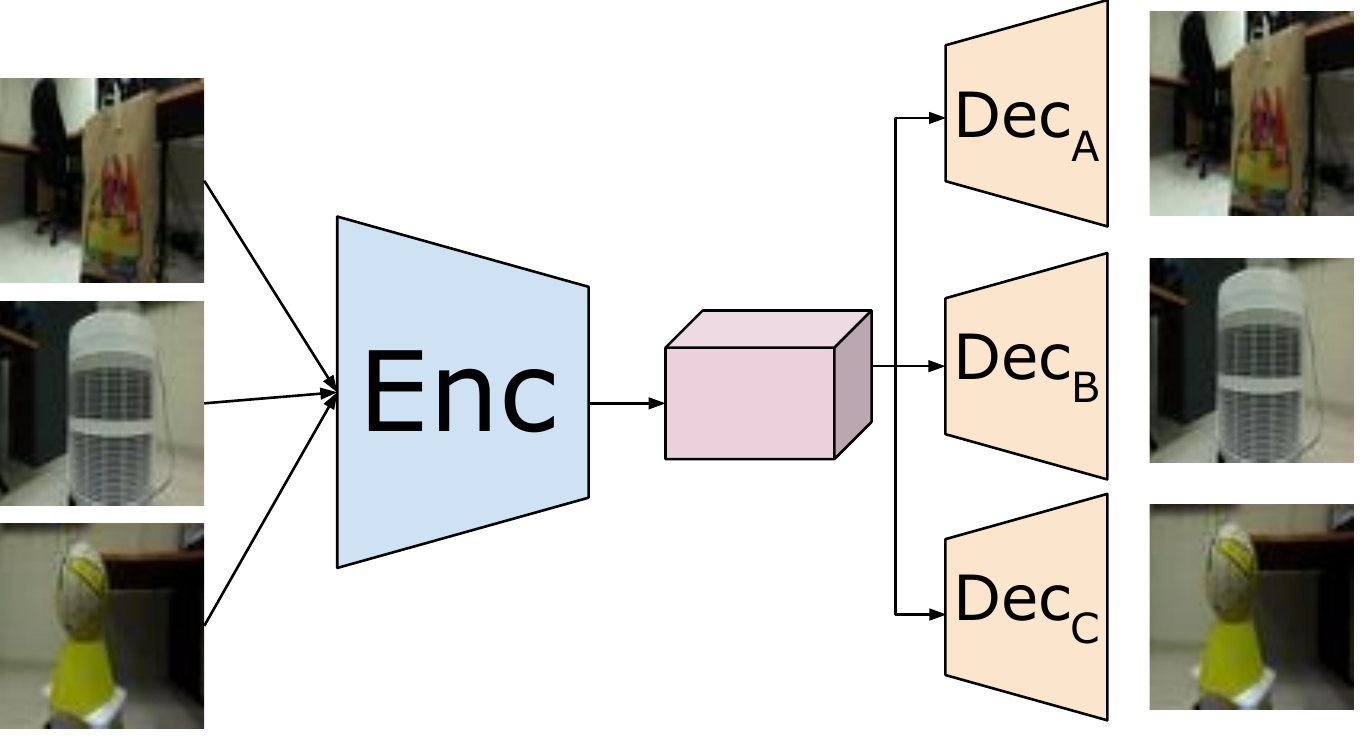}
      \caption{Shared encoder with target specific decoders (TSD). This architecture forces the learned state representation to contain no information specific to the visible target.}
      \label{fig:zeroshot-model}
\end{figure}

\paragraph{Learning shared state-representation} 
In a scenario where we have many different target objects but want to learn a generic policy applicable to all the objects, we can extend our dreaming model by training an autoencoder using target specific decoders (TSD).
Using a single decoder (shared by all images/objects) requires the state representation to maintain certain information about object appearances or types, in order for the decoder to reconstruct it. On the other hand, if we use a target specific decoder, we are able to encourage the representation to contain no object-dependent information, making the target-specific-decoder responsible for reconstructing the object appearance (Fig.~\ref{fig:zeroshot-model}). Having such encoder allows for better policy learning as the policy does not depend on which object is present in our scenarios. This has the effect of reducing the size of the state space. Table~\ref{tab:norml1} confirms that this representation benefits the overall learning.



\begin{table}
      \caption{Normalized $L_1$ distances (Eq.~\ref{eq:rel}) between actual vs. predicted future state.}
      \label{tab:norml1}
      \centering
      \small
      \begin{tabular}{lcc}
        \toprule
             &  Relative ($D_1$)    & Mean ($D_2$) \\
        \midrule
         Standard-ImageNet &	0.98 &	0.741\\
         Standard-Reward &	0.97 &	0.729\\
         E2C \cite{watter2015embed} & 0.95 & 0.82\\
         Dreaming (linear) &	0.82 &	0.558\\
         Dreaming (conv) &	0.71 &	0.448\\
         Dreaming (TSD)         & 0.68  & 0.412\\
        \bottomrule
       \end{tabular}
\end{table}

\begin{figure}
    \centering
      \includegraphics[width=0.48\textwidth]{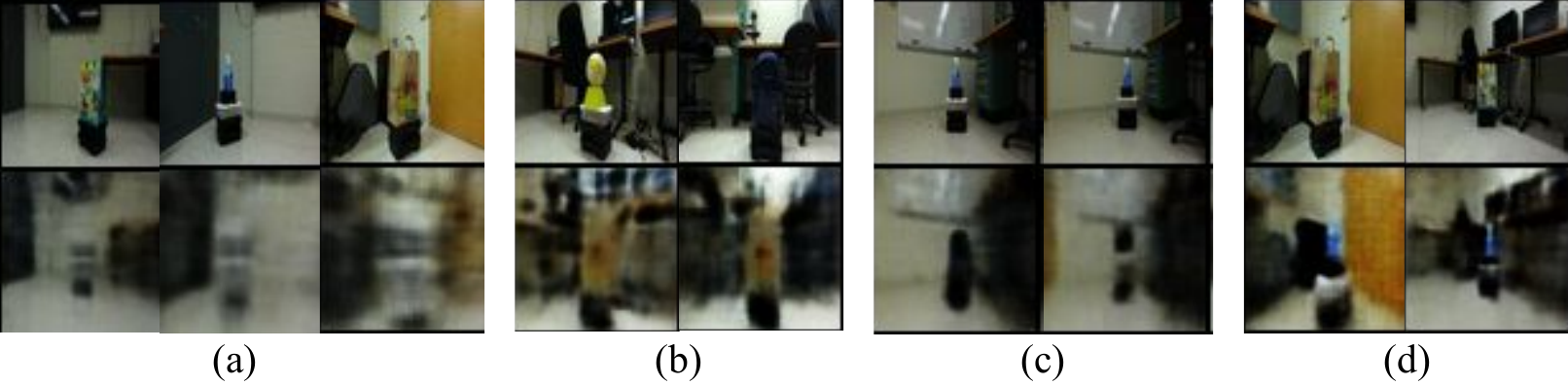}
      \caption{Example of reconstructing unseen targets. Turning \textbf{(a)} bags and a bottle into an airfilter; \textbf{(b)} a volleyball and a backpack into a bag; \textbf{(c)} a bottle into a backpack; and \textbf{(d)} a bag into a bottle.}
      \label{fig:zeroshot-example}
\end{figure}

Additionally, the use of such architecture allows us to recognize unseen target objects. We can apply the encoder to an image containing an unseen target object, then apply one of the decoders to the representation to reconstruct the scene with the other target. Examples of this are shown in Fig.~\ref{fig:zeroshot-example}. We confirm this further in Subsection \ref{sec:real} where we force the robot to interact with unseen objects.


\subsection{Robot tasks and policy learning}
Using our trained dreaming model, we are able to learn different policies entirely in the dreamed environment. We conduct the experiments with two different robot tasks. In one task, (i) we learn a policy to `approach the target object', $\pi_{ap}$. We train a CNN, $H(s_t)$ to output the binary classification if the given state is a goal state. 
For this task, our reward is -1 for all actions taken and +100 when the robot reaches the target. That is, $R(s_t) = 100$ if $H(s_t) > 0.5$ and -1 otherwise.
 This causes the agent to learn a policy to reach the target in as few steps as possible.
 In Fig.~\ref{fig:rand-dream}, we show an example dream sequence starting from a randomly sampled (i.e., not real) state and in Fig.~\ref{fig:real-v-dream} we show a real sequence compared to the dreamed version with the same starting state and actions taken. These figures show that our dreaming model generates realistic trajectories that match the real-world dynamics.

Using the same dreaming model, (ii) we also learn a policy to `avoid the target', $\pi_{avoid}$. Our reward function is $\Delta v$ for each action and $R(s_t) = -100$ when $H(s_t) > 0.5$. This makes the agent learn to move as much as possible, without hitting the target.


\subsection{Offline evaluation (using real-world robot data)}
To evaluate various models and learned policies without running real-world robot trials, we collected a dataset consisting of image frames from the robot's point of view, together with the ground truth geolocations of the robot and the target object.
Given one image (from the real robot trajectory) as the start state, the task is to generate a dreamed trajectory from it by sequentially selecting actions following the learned policy while relying on the state-transition model.
We ran this for 30 steps or until the goal detector CNN predicted the target had been reached. We then computed the distance between the ground truth location of the target object and the robot location after taking such $\sim30$ actions. If the distance is within 20cm of the target, we considered it a successful trial. Note that a successful trial happens only when both the dreaming model and the reinforcement policy learning on top of it are very accurate, since the state transitions have to happen $\sim30$ times. For the avoid task, we force the robot to move at least 20cm each time. If it is within 20cm of the target, we consider it a failed trial. If after 30 actions it has not reached the target, we consider it a successful trial and measure the distance from the target.



Table~\ref{tab:off} compares accuracies of several different reinforcement learning algorithms with baselines and dreaming models. We explicitly confirm that learning in a dream can be effectively done with various standard policy gradient algorithms (e.g., REINFORCE, Actor-Critic), and genetic algorithms (e.g., CMA-ES). We find that starting from random states in addition to real states leads to a more accurate policy than starting only from real states (i.e., previous methods with imagined-trajectories), confirming that the ability to `dream' is beneficial. We also compare to a baseline policy of constantly moving straight. We also observe that `dreaming' from random, unseen states is not beneficial, sometimes even harmful, when we do not have a realistic representation and state-transition model (e.g., the standard and linear models). However, when provided a realistic state-transition model and representation (e.g., conv model), the ability to dream from unseen, random states is beneficial.

{\setlength{\tabcolsep}{0.4em}
\begin{table*}
\small
  \caption{Offline evaluation of the target `approach' task. We test our dreaming models using various reinforcement learning algorithms, averaged over 500 trials. Note that all RL algorithms only interact with the dreaming model. `Random' means that we use random start states for the dreaming, and `real' means the dreaming uses real states as its start states (i.e., $s_0$).}
  \label{tab:off}
  \centering
  \begin{tabular}{lccccccccc}
    \toprule
     & \multicolumn{3}{c}{Standard-Reward} &    \multicolumn{3}{c}{Dreaming (linear)} & \multicolumn{3}{c}{Dreaming (conv)}               \\
    \cmidrule(r){2-4}  \cmidrule(r){5-7} \cmidrule(r){8-10}
         &  Random    & Real & Combined &   Random    & Real & Combined  &  Random    & Real & Combined  \\
    \midrule
    Constant Forward & 22\%  \\
    MPC \cite{nagabandi2017neural}            & - & 24\% & -  & - & 28\% & - & - & 28\% & - \\
    REINFORCE      & 25\% & 26\% & 26\% & 26\% & 44\% & 39\% & 54\% & 50\% & 55\% \\
    Actor-Critic   & 25\% & 26\% & 26\% & 34\% & 34\% & 34\% & 55\% & 52\% & 61\% \\
    CMA-ES        & 26\% & 27\% & 25\% & 32\% & 36\% & 35\% & 57\% & 54\% & 62\% \\
    \bottomrule
  \end{tabular}
\end{table*}
}

\begin{table}
\scriptsize
  \caption{Offline evaluation of the target approach/avoid tasks. We report the percentage of successful trials, based on the policies learned with Actor-Critic. We also measure the robot's average distance from the target in the avoidance task. Unlike other methods, E2C requires explicit end states to be provided as its input, and thus was not applicable for the avoid task.}
  \label{tab:off-compare}
  \centering
  \begin{tabular}{lcc}
    \toprule
         &  Approach & Avoid (Dist) \\
    \midrule
     Standard-ImageNet  & 25\% & 1.1m \\
     Standard-Reward  & 26\%  & 1.1m \\
     E2C + control \cite{watter2015embed} & 25\% & - \\
     E2C state-transition & 26\% & - \\
     Dreaming (linear) & 39\% & 1.4m \\
     Dreaming (conv) & 62\% & 1.8m\\
     Dreaming (TSD)  & 65\% & 2.1m\\
    \bottomrule
  \end{tabular}
\end{table}

\begin{figure}
    \centering
      \includegraphics[width=0.42\textwidth]{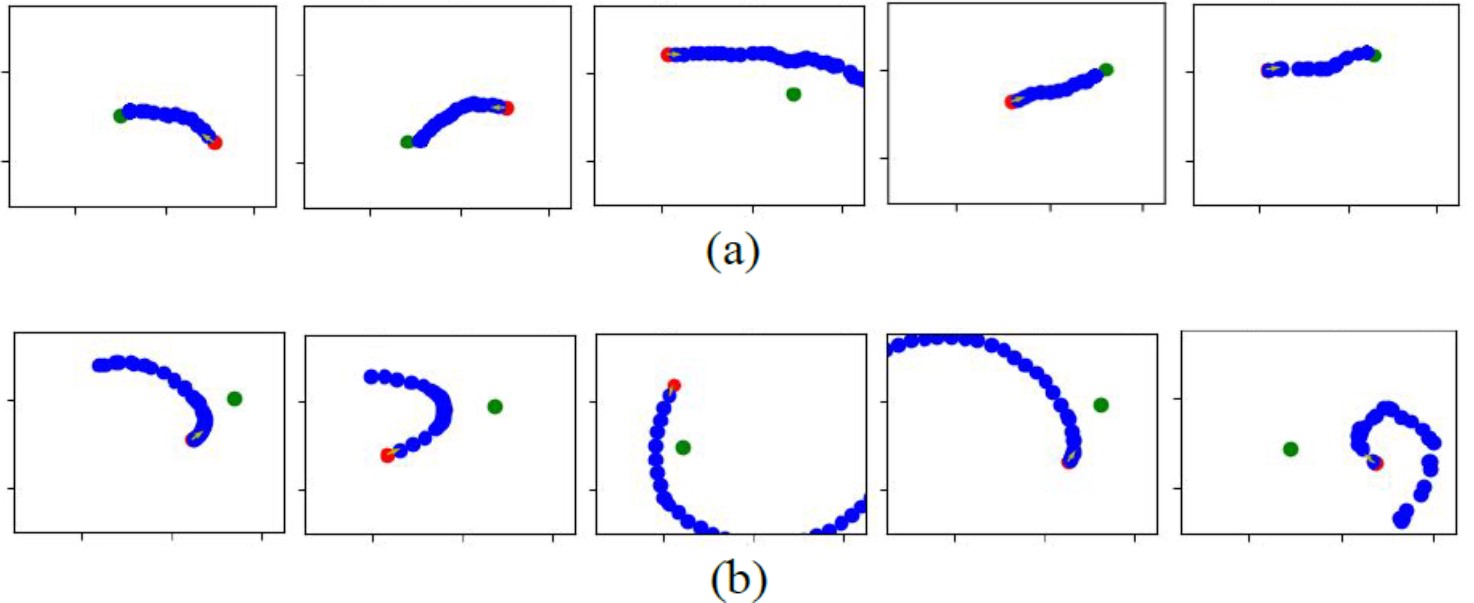}
    \caption{The red dot represents the robot's starting location, the green dot is the target location, the blue dots are the resulting locations of the actions. On average, robot starts 2 meters away from the target. We show offline trajectories for the \textbf{(a)} approach task and \textbf{(b)} avoid task.}
    \label{fig:ex-traj}
\end{figure}

In Table \ref{tab:off-compare} we compare the learned policies using Actor-Critic with baselines and dreaming models for both the target approach and the avoid tasks. To train these policies, we use both real and random starting states. We find that the standard state representation/transition model learning result in a constant policy. The linear representation trained with the decoder performs better, but not as well as our dreaming with convolutional representations. This confirms that maintaining the spatial information is beneficial for policy learning. Additionally, we find that using a single encoder to learn a shared representation for all target objects provides the best results. Fig.~\ref{fig:ex-traj} shows examples.






\subsection{Real-world robot experiments}
\label{sec:real}

\begin{figure*}
    \centering
    \includegraphics[width=\textwidth, height=1.7cm]{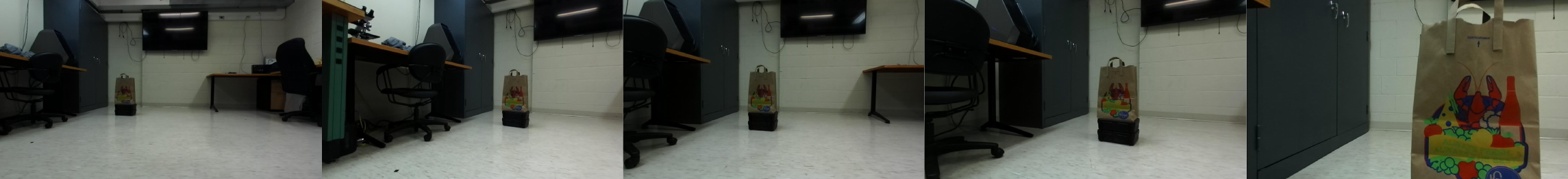}
      \includegraphics[width=0.65\textwidth]{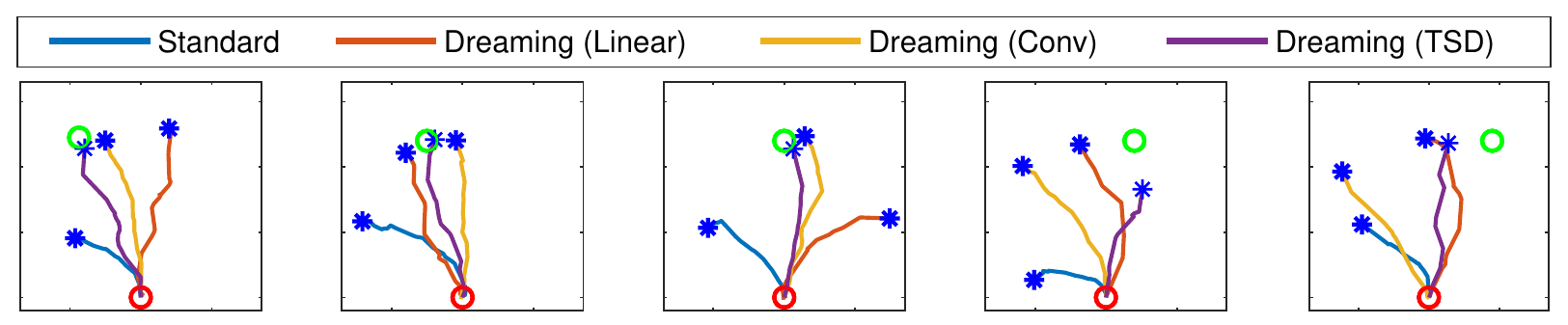}\rulesep
      \includegraphics[width=0.25\textwidth]{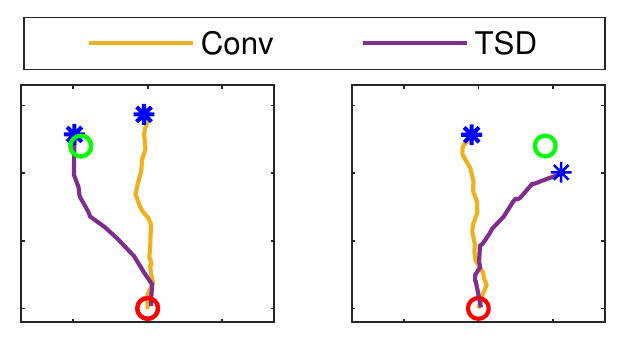}
      \caption{Example real-world robot frames (top), and trajectories of different models during the `approach' task for seen (bottom left) and unseen (bottom right) targets. The trajectories shown are just for the visualization, and the robot generated its actions solely based on the RGB inputs without any information on localization or target detection.} 
      \label{fig:real-traj}
      \vspace{-12pt}
\end{figure*}

In Table~\ref{tab:real}, we compare the performance of various models trained entirely in a dream and directly applied to the real-world setting. We demonstrate the ability to learn different policies for two different tasks (i.e., approaching and avoiding), using a single learned state-transition model, without any real-world robot trials. Actor-Critic was used as our reinforcement learning algorithm in this experiment. We are able to confirm that our dreaming model obtains better success rates on both tasks compared to the standard approach of using CNN states. Our convolutional dreaming showed improved accuracy. Additionally, we conducted an experiment to approach/avoid an `unseen' target. We find that our TSD approach allows the RL agent to learn a general policy, successfully reacting to different targets. Fig.~\ref{fig:real-traj} shows a sequence of images the robot received during the approach task. Once learned, our action policy CNN runs in real-time on a Nvidia Jetson TX2 mobile GPU. 




\begin{table}
\small
  \caption{Accuracies measured with real-world robot trials on both tasks. We tested 15 total trials split between 3 targets for each model in the approach task and 10 trials with 2 objects in the avoidance task. The target was on average 2.5 meters away from the robot for the approach target task and 0.6 meters away in the avoid task. We report the percentage of successful trials. All approaches were trained solely based on the dreaming models without any real-world interactions (except for initial random action samples). Most of the baselines failed in unseen target approach/avoid tasks.}
  \label{tab:real}
  \centering
  \begin{tabular}{lcccc}
    \toprule
       & \multicolumn{2}{c}{Seen Targets} &    \multicolumn{2}{c}{Unseen Targets}\\
       \cmidrule(r){2-3}  \cmidrule(r){4-5}
         &  Approach    & Avoid  & Approach & Avoid\\
    \midrule
     Standard-Reward & 0\% & 50\%  & - & -\\
     E2C \cite{watter2015embed} & 20\% & - & - & -\\
     Dreaming (linear) & 27\% & 40\% & - & - \\
     Dreaming (conv) & 47\% & 100\% & 40\% & 20\%\\
     Dreaming (TSD) & 60\% & 100\% & 60\% & 90\%\\
    \bottomrule
  \end{tabular}
\end{table}



\subsection{Evaluation on different environment/task}

\begin{figure}
    \centering
    \includegraphics[width=3.8cm, height=2.4cm]{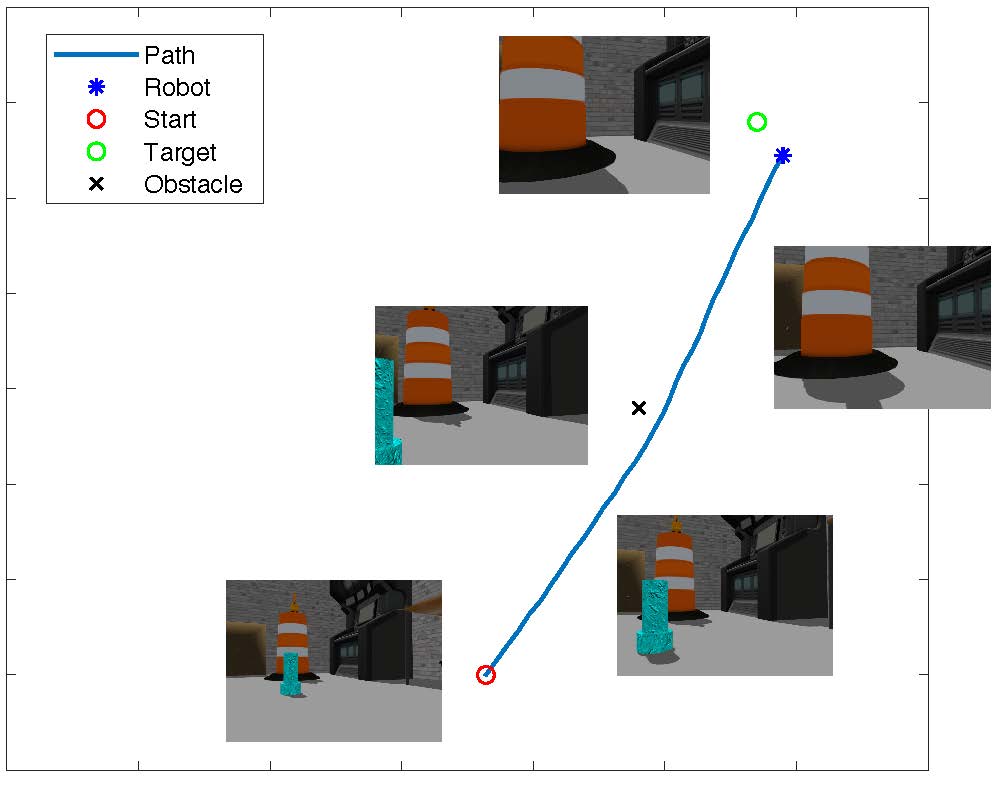}
    \includegraphics[width=0.22\textwidth, height=2.4cm]{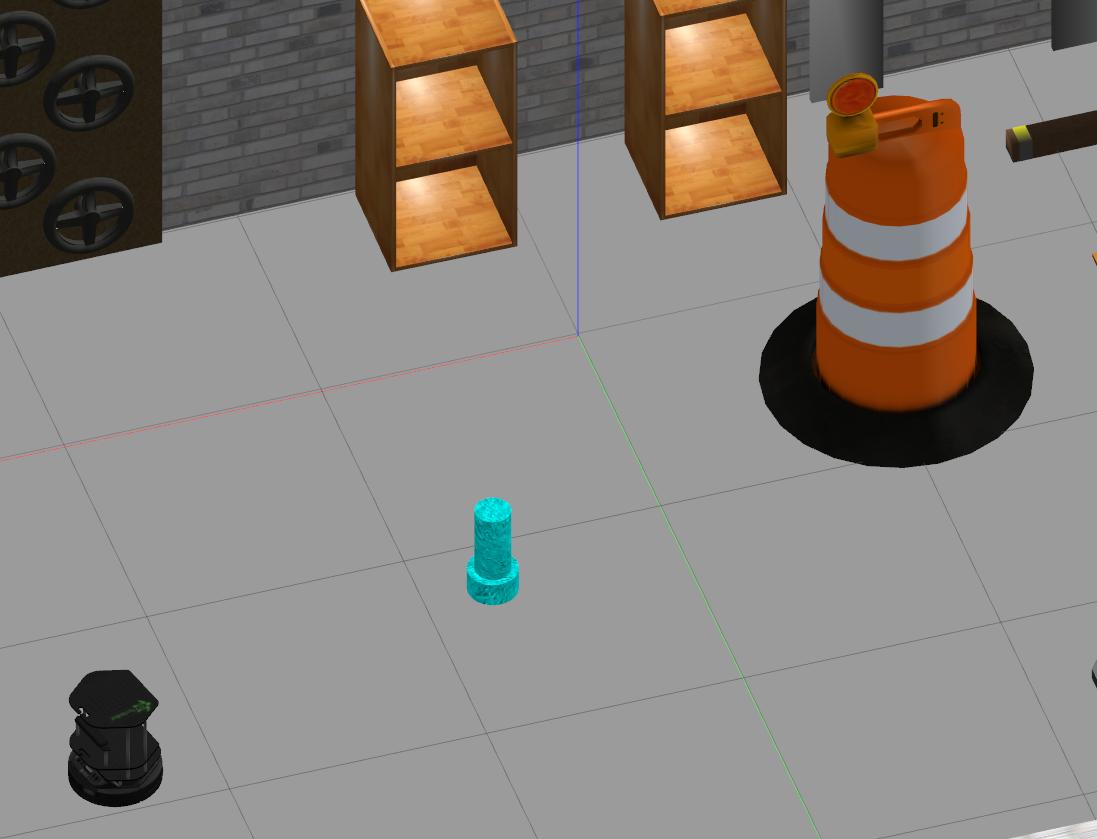}
      \caption{An example trajectory with robot's view (left) and visualization of the simulated environment with the robot (right) used in our obstacle experiment.  The robot's task is to approach the target (construction barrel) while successfully avoiding an obstacle (standing pipe).}
      \label{fig:sim_traj_single}
      \vspace{-16pt}
\end{figure}

We also confirm the effectiveness of our approach in a different environment with a more challenging task. Fig. \ref{fig:sim_traj_single} illustrates example frames from an environment constructed with the Gazebo simulator. 
We placed an obstacle between the robot and target object which the robot had to learn to avoid while moving, making the task of approaching the target more challenging. Note that the view of the target is partially obstructed by the obstacle at various robot locations.

In Table \ref{tab:off-compare-sim}, we show the robot task success rate measured in this simulated environment over 16 trials, compared with other approaches. As in our original setting, only a set of initial samples were provided for the robot to learn the dreaming model and the robot had to solely rely on it for the policy learning. The experimental results confirm that our exact same approach generalizes well to a completely different environment and is applicable for more complicated tasks.

Here, we also compare to a more traditional scene representation for this task by using an object detection-based occupancy grid as the state representation. This was done by using bounding boxes based on the SSD \cite{liu2016ssd} object detector trained on MS COCO. The state representation had a size of 32x32x80, where 80 is the number of channels corresponding to MS COCO classes. As we can confirm in Table~\ref{tab:off-compare-sim}, this (together with the learned state-transition model) does provide more reasonable baseline performance, but is still worse than our proposed method.

\begin{table}
\scriptsize
  \caption{Evaluation of the target approach task while avoiding an obstacle. We report the percentage of successful trials, based on the policies learned with Actor-Critic.}
  \label{tab:off-compare-sim}
  \centering
  \begin{tabular}{lccc}
    \toprule
         &  Success Rate (Online) & Offline\\
    \midrule
     Standard-ImageNet  & 6\% & 25\%\\
     E2C & 19\% & 12\%\\
     Object occupancy-based state (SSD)  & 38\% & 50\%\\
     Dreaming (conv) & 75\% & 70\%\\
    \bottomrule
  \end{tabular}
\end{table}

\section{Conclusion}

We proposed learning real-world robot action policies by dreaming. The main idea is to learn a realistic dreaming model that can emulate samples equivalent to frame sequences from the real environment, and have the robot learn policies by interacting with the dreaming model. 
We experimentally confirmed that we are able to learn a realistic dreaming model from only a few initial random action samples, and showed that the dreaming model together with a reinforcement learning algorithm enable learning of visuomotor policies directly applicable to real-world robots.

{\ssmall
\bibliographystyle{IEEEtran}
\bibliography{bib}
}

\newpage

\appendix
\section{Implementation details}
\subsection{Model architectures}
We implement our models in PyTorch. Our convolutional autoencoder takes images of size $64\times 64$ as input. The encoder network has 4 convolutional layers, all with kernel size $4\times 4$ with a stride of 2, no padding and followed by a ReLU activation function. The layers have 32, 64, 128, and 256 channels. We then have two $1\times 1$ convolutional layers with no activation function. These layers produce $\mu$ and $\sigma$ used to obtain the latent representation.  This results in a $2\times 2\times 256$ dimensional state representation

The decoder network starts with a convolutional layer with 256 channels and a $3\times 3$ kernel, stride of 1 and padding of 1 followed by a ReLU activation function. The decoder then has 5 transposed convolutional layers (also called deconvolutional layers) with kernel size $4 \times 4$ and upsampled by a factor of 2. The layers have 128, 64, 32, and 3 channels. This results in the output being a $64\times 64$ image.

The action representation network has 2 fully connected layers, going from 3 to 64 to 128. Since our actions can be negative, we use the LeakyReLU activation function:
\begin{equation}
    \text{LeakyReLU}(x)\begin{cases} 
      -0.2x & x\le 0 \\
      x & 0\geq 0
   \end{cases}
\end{equation}
The 128-dimensional vector is reshaped into a $2\times 2\times 32$ tensor. This is used as input to a convolutional layer with a $3\times 3$ kernel with 32 channels, stride of 1 and padding of 1, followed by the LeakyReLU function. The next layer is a $1\times 1$ convolutional layer with 64 channels followed by the LeakyReLU.

The output of the action representation network is concatenated with the state representation giving a $2\times 2\times 256+64$ dimensional representation. This is used as input to the future regression (or state-transition) CNN. This CNN consists of 4 convolutional layers, each followed by the LeakyReLU function.  The first layer has 512 channels, a $3\times 3$  kernel with stride of 1 and padding of 1. The second has 512 channels,  $1\times 1$ kernel, a stride of 1 and no padding. The third layer has 256 channels a $3\times 3$ kernel and padding of 1. The final layer has 256 channels and a $1\times 1$ kernel.

The reward/end detector CNN has 2 convolutional layers. The first is $2\times 2$ with 32 channels, followed by a ReLU activation function and the second is $1\times 1$ with 1 channel followed by a sigmoid activation function. This network predicts the probability of a given $2\times 2\times 256$ state representation being the goal.

The actor/policy network have the same architecture regardless of reinforcement learning algorithm. They consist of a convolutional layer with a $2\times 2$ kernel, ReLU, convolutional with $1\times 1$ and 3 channels. This produces the action. The critic follows the same architecture, except the final layer has 1 output channel.

\subsection{Training details}

\paragraph{Dreaming model:}
We jointly train the autoencoder and future regressor with the Adam method. We set the learning rate to 0.001 for 20 epochs. After every 10 epochs, we decay the learning rate by a factor of 10. 

\paragraph{Reinforcement learning:}
To train the actor-critic and REINFORCE networks, for each iteration we use a batch of 256 trajectories run up to 30 steps. These trajectories are obtained from our dreaming model and was not from the real-world environment; we are able to generate as many trajectories as we want for any policy as needed. Note that this takes very little time to execute as only the future regressor and reward CNNs run all on a GPU. We use the Adam method with the learning rate set to 0.01 for 2000 iterations.

To train policies with CMA-ES, we uses a population size of 128. For each candidate network, we used a batch of 256 trajectories and ran each for up to 30 steps. We computed the mean reward for the batch and followed the CMA-ES algorithm to generate the next candidate networks. We ran CMA-ES for 50 iterations.

\section{Additional examples}
\begin{figure*}
    \centering
      \includegraphics[width=\textwidth]{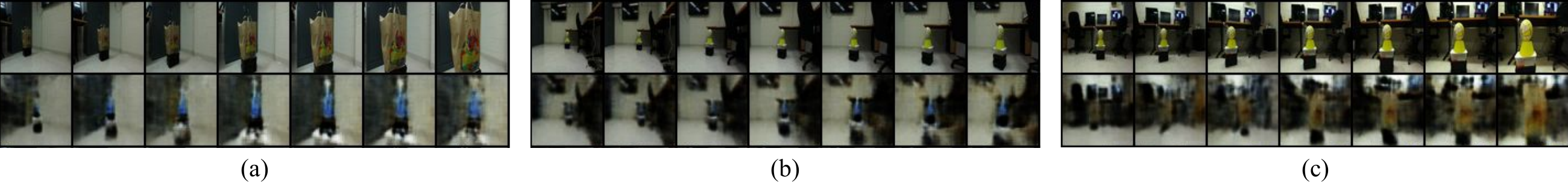}
      \caption{Example sequence of target transfer. \textbf{(a)} A sequence of an unseen bag transferred into a bottle. \textbf{(b)} An unseen volleyball transferred into a bottle. \textbf{(c)} An unseen volleyball turned into a bag.}
      \label{fig:zeroshot-example-seq}
\end{figure*}

In Fig~\ref{fig:zeroshot-example-seq}, we show three example sequences of the target transfers. The model is given input of an image containing an unseen target. Applying one of the target-specific decoders produces an image with the other target.

\begin{figure*}
    \centering
      \includegraphics[width=\textwidth]{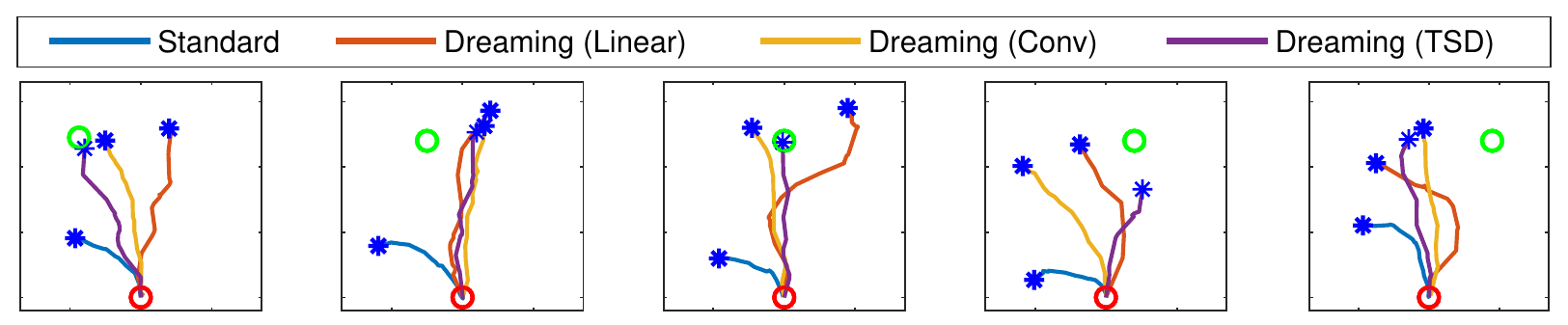}
      \includegraphics[width=\textwidth]{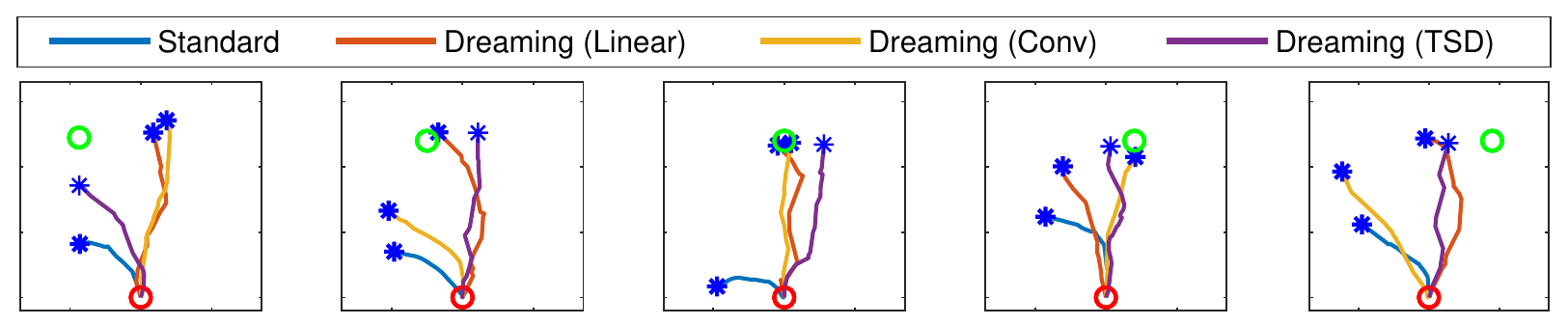}
      \caption{Trajectories on the approaching task taken by the robot in the real world for various different seen targets. The target was on average 2.5 meters away from the robot.}
      \label{fig:real-traj-add}
\end{figure*}

\begin{figure*}
    \centering
      \includegraphics[width=\textwidth]{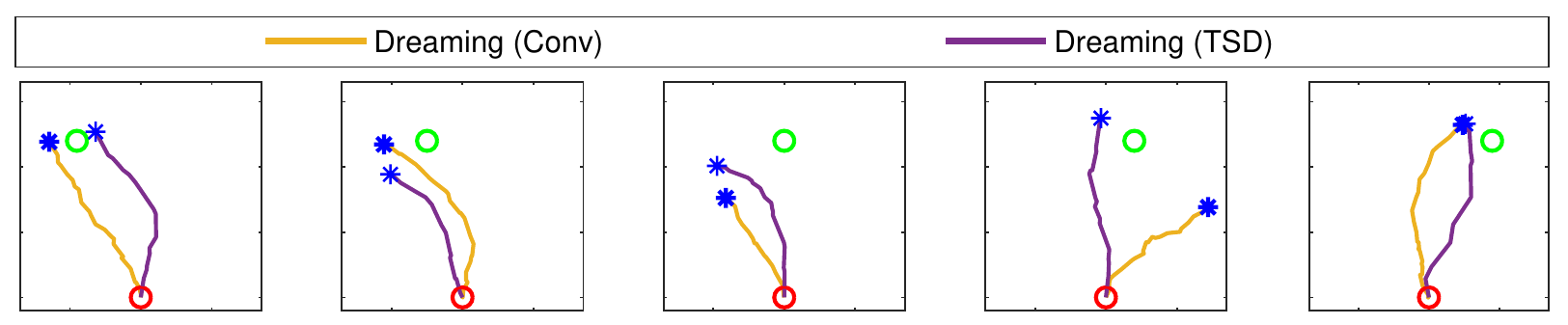}
      \includegraphics[width=\textwidth]{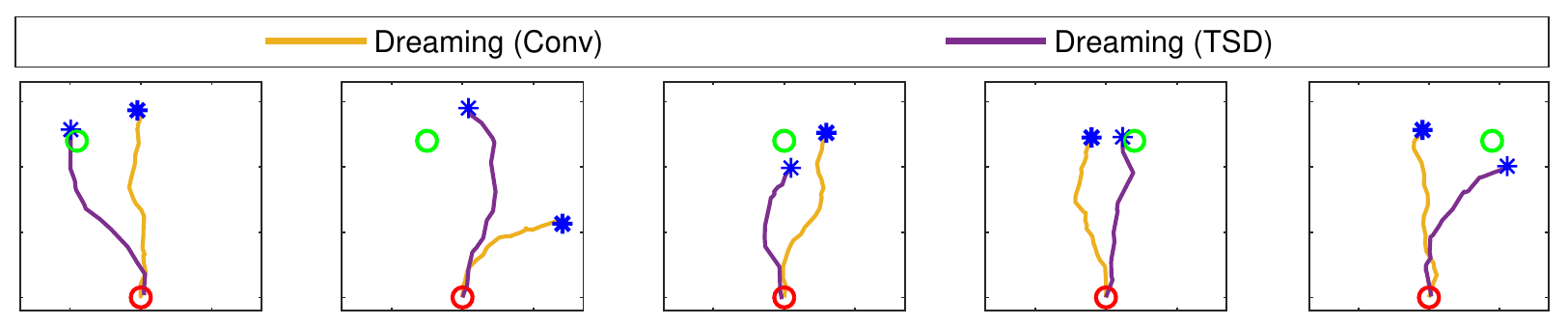}
      \caption{Trajectories on the approaching task taken by the robot in the real world for unseen targets.}
      \label{fig:real-traj-seek-zero}
\end{figure*}

\begin{figure*}
    \centering
      \includegraphics[width=0.6\textwidth]{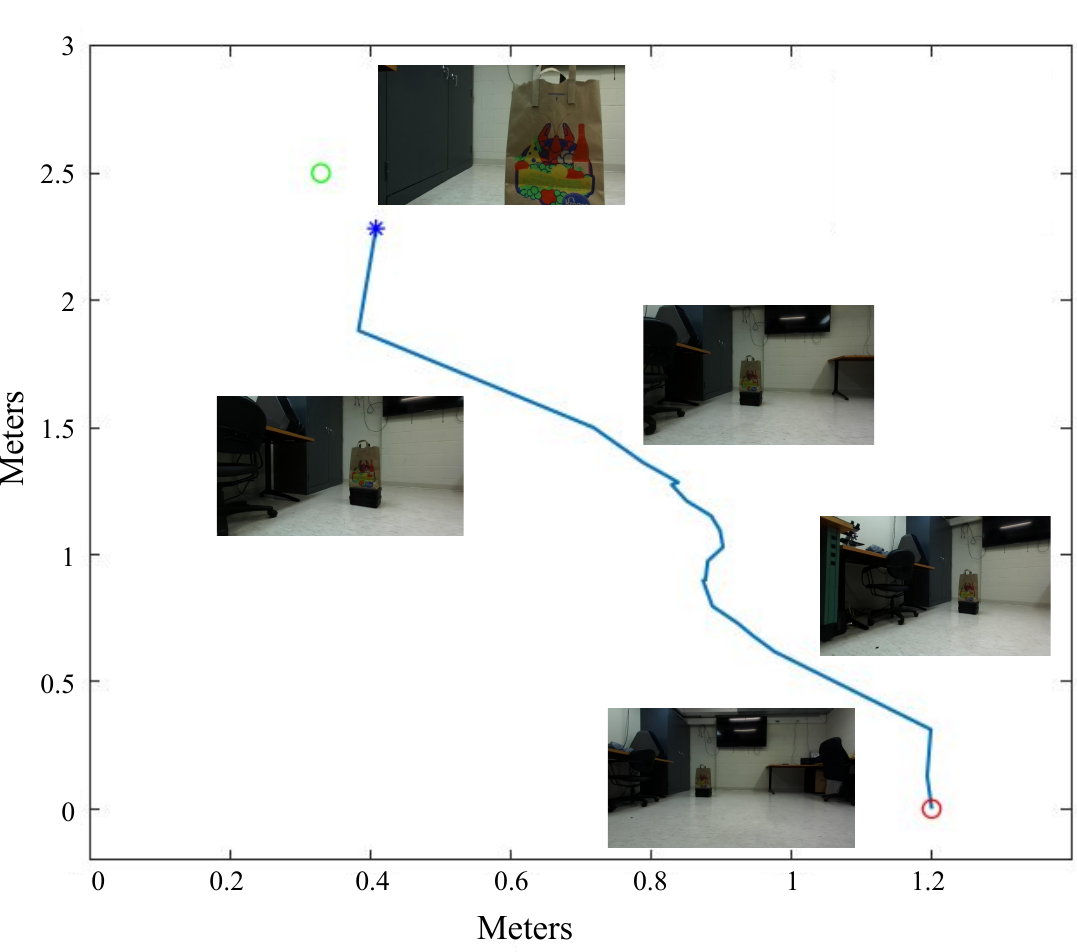}
      \caption{A real-world trajectory annotated with the robot's point of view.}
      \label{fig:real-loc}
\end{figure*}

In Fig.~\ref{fig:real-traj-add}, we show example real-world robot trajectories for approaching seen targets. In Fig.~\ref{fig:real-traj-seek-zero}, we shown example real-world trajectories for approaching unseen targets. In Fig.~\ref{fig:real-loc}, we show an example trajectory annotated with the robot's point of view.

\begin{figure*}
    \centering
      \includegraphics[width=\textwidth]{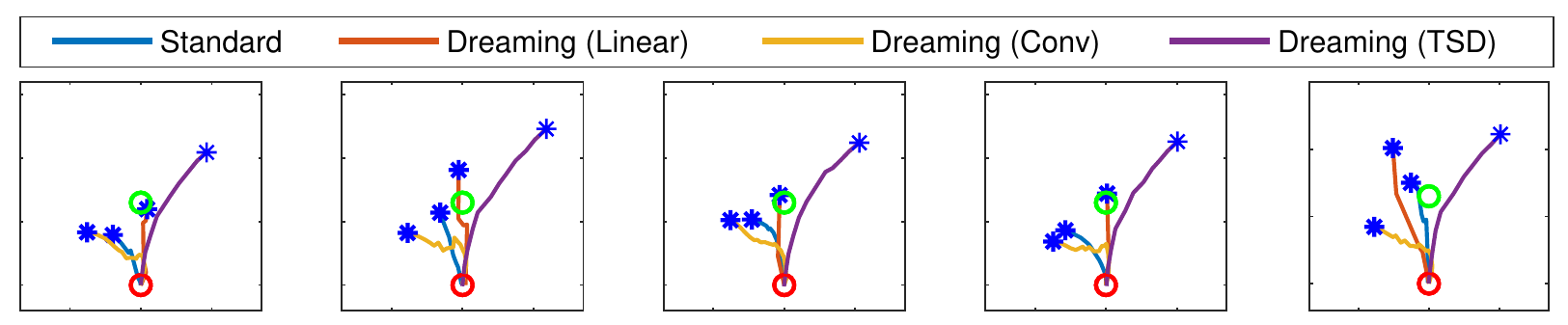}
      \includegraphics[width=\textwidth]{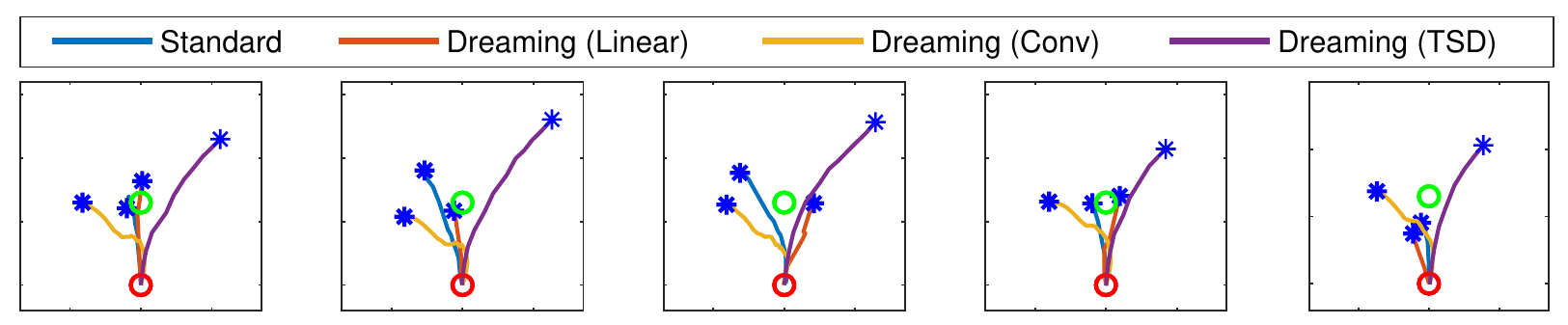}
      \caption{Trajectories on the avoiding task taken by the robot in the real world for various seen targets. To make this task more challenging, the target was placed on average 0.6 meters away from the robot. }
      \label{fig:real-traj-avoid}
\end{figure*}

\begin{figure*}
    \centering
      \includegraphics[width=\textwidth]{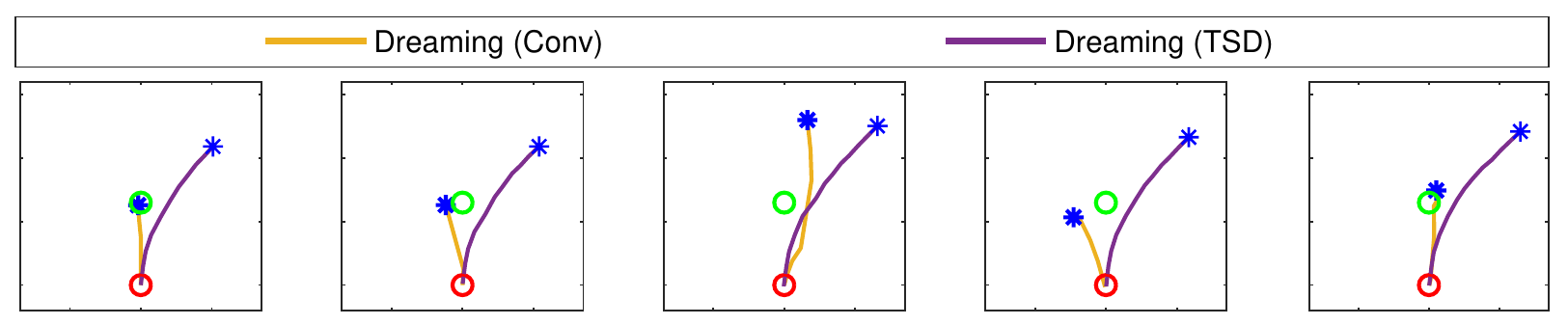}
      \includegraphics[width=\textwidth]{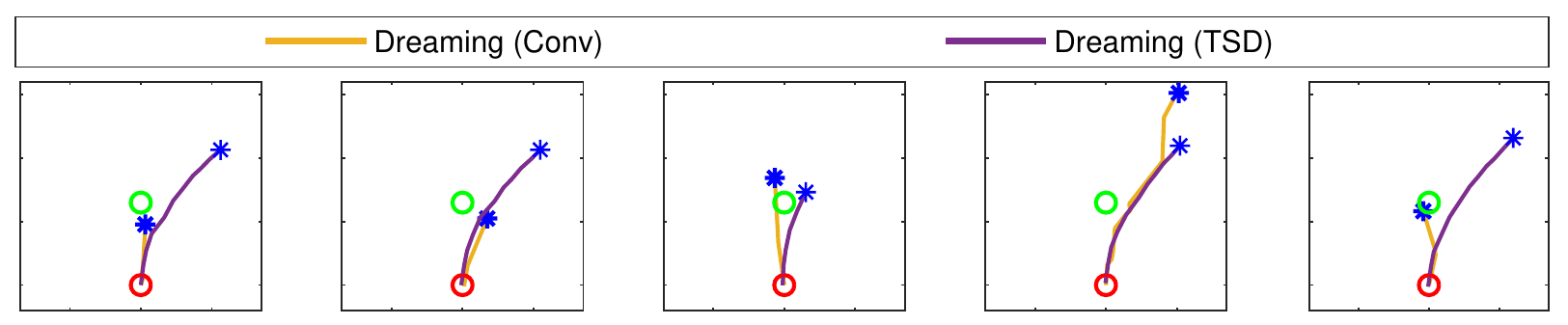}
      \caption{Trajectories on the avoiding task taken by the robot in the real world for unseen targets. To make this task more challenging, the target was placed on average 0.6 meters away from the robot. }
      \label{fig:real-traj-avoid-zero}
\end{figure*}

In Fig.~\ref{fig:real-traj-avoid}, we show several real-world robot trajectories for avoiding a seen target. In Fig.~\ref{fig:real-traj-avoid-zero}, we show several example trajectories for avoiding an unseen target.

\section{Dreaming Model Studies}
We experimentally compared many different models to find the one best suited for dreaming. In this section, we show results for various amounts of initial random samples, various input image and latent representation resolutions and comparison to standard online reinforcement learning using simulated environments.

\subsection{Number of initial random samples}
We collected 400 random trajectories which were on average 100 actions long, resulting in about 40,000 state-action pairs. To determine how many initial random samples were needed, in Table \ref{tab:num-samples}, we compare our TSD model trained with various amounts of data. We report both the mean $L_1$ distance (Eq. 5) and the offline evaluation for the approach task (Section 6.4). We find that  5,000 action samples (50 trajectories) are sufficient to learn a good dreaming model for this task.

\begin{table}
\small
  \caption{Evaluation of the target approach task with different amounts of initial random samples (i.e., training data). Here we used the TSD model. We report both the $L_1$ difference (Eq. 5) and the offline evaluation (Section 6.4) for the approach task.}
  \label{tab:num-samples}
  \centering
  \begin{tabular}{lcc}
    \toprule
         &  Mean (Eq. 5, $D_2$)  &  Approach\\
    \midrule
    1,000 actions & 0.57 & 56\%\\
    2,000 actions & 0.52 & 58\%\\
    5,000 actions & 0.49 &  63\%\\
    10,000 actions & 0.45 & 64\%\\
    20,000 actions & 0.41 & 65\%\\ 
    40,000 actions & 0.40 & 65\%\\ 
    \bottomrule
  \end{tabular}
\end{table}

\subsection{Resolution of input and representation}
To determine the effect of spatial information, we compare several different TSD dreaming models with various input image resolutions and different latent representation sizes, shown in Table \ref{tab:inp-res}. We find that using $64\times 64$ input image with our chosen architecture performed the best. We found that with the used encoder-decoder architecture (Appendix A), larger images resulted in worse reconstructions. Without good reconstructed images, the state-transition and dreaming models failed, which lead to poor policy learning. 

In Table \ref{tab:rep-res}, we compared different sizes of the representation. Here, we found that a $2\times 2$ representation performed the best. This representation maintains some spatial information, which is important for reconstruction and future prediction, while the larger representations lead to overfitting and poor policy learning.

\begin{table}
  \caption{Offline evaluation of different input resolutions and representation sizes using the offline evaluation (Section 6.4) for the approach task.}
  \label{tab:res-exp}
\centering
\begin{minipage}{0.48\textwidth}
  \caption{Evaluation of input sizes}
  \label{tab:inp-res}
  \centering
  \begin{tabular}{lc}
    \toprule
         &  Success\\
    \midrule
    $32\times 32$  & 55\%\\
    $64\times 64$  & 65\%\\
    $128\times 128$  & 59\%\\
    $256\times 256$  & 48\%\\
    \bottomrule
  \end{tabular}
\end{minipage}
\hfill
\begin{minipage}{0.48\textwidth}
  \caption{Evaluation of representation sizes on the approach task.}
  \label{tab:rep-res}
  \centering
  \begin{tabular}{lc}
    \toprule
         &  Success\\
    \midrule
    $1\times 1$  & 39\%\\
    $2\times 2$  & 65\%\\
    $4\times 4$  & 60\%\\
    $8\times 8$  & 42\%\\
    \bottomrule
  \end{tabular}
\end{minipage}
\end{table}

We also compare a version of our model that does not use the convolutional action representation. We compare two different methods: (i) using fully-connected layers to create a 32-dimensional action representation which are concatenated to each spatial location and (ii) as before, create a 128-dimensional vector which is reshaped into a $2\times 2\times 32$ tensor which is then concatenated with the state representation (no convolutional layers used). The results, shown in Table \ref{tab:conv-act}, confirm that using a spatial representation is better than the linear action and that the convolutional representation is further beneficial.

\begin{table}
  \caption{Offline evaluation of action representations on the approach task.}
  \label{tab:conv-act}
  \centering
  \begin{tabular}{lcc}
    \toprule
         & Mean (Eq. 5, $D_2$)  & Success\\
    \midrule
    Linear action representation  & 0.52 & 48\%\\
    Spatial action representation  & 0.48  & 57\%\\
    Conv action representation  & 0.41 & 65\%\\
    \bottomrule
  \end{tabular}
\end{table}

\end{document}